\documentclass{article}
\usepackage{kr}

\usepackage{times}
\usepackage{soul}
\usepackage{url}
\usepackage[hidelinks]{hyperref}
\usepackage[utf8]{inputenc}
\usepackage[small]{caption}
\usepackage{graphicx}
\usepackage{amsmath}
\usepackage{amsthm}
\usepackage{booktabs}
\usepackage{algorithm}
\usepackage{algorithmic}
\usepackage{caption}
\usepackage{subcaption}
\usepackage{lipsum}
\usepackage{rotating}

\newtheorem{example}{Example}
\newtheorem{theorem}{Theorem}
\newtheorem{lemma}{Lemma}
\newtheorem{definition}{Definition}

\title{Propositional Encodings of Acyclicity and Reachability by using Vertex Elimination}

\author{%
Masood Feyzbakhsh Rankooh$^1$\and
Jussi Rintanen$^1$ \\
\affiliations
$^1$ Department of Computer Science\\
Aalto University, Helsinki, Finland\\
\emails
masood.feyzbakhshrankooh@aalto.fi,
jussi.rintanen@aalto.fi
}

\begin{document}

\maketitle

\begin{abstract}
We introduce novel methods for encoding acyclicity and s-t-reachability constraints for propositional formulas with underlying directed graphs. They are based on \emph{vertex elimination graphs}, which makes them suitable for cases where the underlying graph is sparse. In contrast to solvers with ad hoc constraint propagators for acyclicity and reachability constraints such as GraphSAT, our methods encode these constraints as standard propositional clauses, making them directly applicable with any SAT solver. An empirical study demonstrates that our methods together with an efficient SAT solver can outperform both earlier encodings of these constraints as well as GraphSAT, particularly when underlying graphs are sparse.
\end{abstract}

\section{Introduction}

Graphs are powerful tools for representing knowledge. Many knowledge representation approaches incorporate graphs to maintain conceptual relations among their elements. Graphs introduce structure to knowledge representation methods. Once such a structure has been assumed, investigating the existence and exploitation of structural properties is only natural. Reachability and acyclicity are two of the most important structural properties of graphs.

Graph constraints are important in knowledge representation languages. For example, acyclicity constraints are part of reductions of Answer Set Programming to SAT \cite{LinZhao04,GebserJR14ecai}, and implicit in fixpoint semantics of inductive definitions \cite{DeneckerTernovska08}. In AI planning, acyclicity is needed in SAT encodings for classical planning that use partial orders \cite{RintanenHN06}, and for non-deterministic and partially observable planning \cite{chat,PandeyRintanen18}. Moreover, constraint-based methods for structure learning of Bayesian networks need the acyclicity of the networks with graph constraints \cite{tc3}. 

The above-mentioned approaches have motivated the development of better encodings of acyclicity and other graph constraints in the propositional logic, as well as the study of specialized propagators for these constraints.

In this work we address the satisfiability of propositional formulas with underlying directed graphs, under reachability and acyclicity constraints. The motivation for our work is the difficult trade-off between size and propagation strength in existing clausal encodings of these constraints \cite{GebserJR20} on one hand, and the effort in implementing specialized graph constraint propagators \cite{graphsat}, and adapting and embedding them in new SAT solvers as ones become available, on the other.

Our goal is to develop encoding methods for graph constraints such as acyclicity and reachability, that are competitive with specialized ad hoc graph constraint propagators, and which suffer less from the large size of those traditional clausal encodings that have good propagation properties. We particularly address sparse graphs.

Our idea is to use vertex elimination graphs \cite{ve:ve} as a structure that preserves reachability and acyclicity properties of the underlying graph, and also allows succinct encoding of graph constraints into propositional formulas, particularly when the underlying graph is sparse. The current state-of-the-art method for satisfying acyclicity and reachability constraints in the SAT context is GraphSAT \cite{graphsat}. While GraphSAT relies on a specialized algorithm for satisfying graph constraints, our methods explicitly encode the constraints into propositional formulas, and therefore allow an easy reuse of the method with any other state-of-the-art SAT solver without additional implementation effort.

We provide the theoretical arguments for correctness of our methods, and also, deliver theoretical evidence for efficiency of the methods by undertaking a parameterized complexity analysis. Moreover, our empirical results show that by employing an efficient SAT solver, our new methods can outperform GraphSAT and other encoding methods, particularly when underlying graphs are sparse.

The rest of this paper is organized as follows. Section 2 provides a formalization of the concepts that are essential for description of our methods. These concepts include propositional formulas with underlying directed graphs and vertex elimination graphs. In Section 3, we survey previously introduced methods for enforcing acyclicity and reachability constraints. Section 4 introduces our novel methods for guaranteeing graph constraints for formulas with underlying directed graphs. We also provide theoretical arguments for correctness and efficiency of our methods in Section 4. In Section 5, we present our empirical results and discuss the potentials and limitations of our methods. Section 6 concludes the paper. 

\section{Preliminaries}
In this section we provide formal definitions for propositional formulas with underlying directed graphs, encoding of graph constraints, and vertex elimination graphs, along with related concepts.
\subsection{Propositional Formulas with Underlying Directed Graphs}
Let $\phi$ be a propositional formula over the set of propositions $X$, and $Y$ be a subset of $X$, such that every member of $Y$ represents an edge of graph $G=(V,E)$. We call $\phi$ a propositional formula with underlying directed graph $G$, and denote the proposition that represents $(v_i,v_j)\in E$ in $\phi$ by $e_{i,j}$. If there exists a model $\mathcal{M}$ for $\phi$, we construct $G_\mathcal{M}=(V,E_\mathcal{M})$, the underlying graph of $\mathcal{M}$, where $E_\mathcal{M}=\{(v_i,v_j)|\mathcal{M}(e_{i,j})=true\}$.

For propositional formulas with underlying directed graphs, one can enforce certain constraints on the underlying graphs by conjunction of the original formulas with additional formulas.
\begin{definition}[Encoding of acyclicity]
Let $\phi$ be a propositional formula with underlying graph $G$. The encoding of acyclicity for $\phi$ is a propositional formula $\phi_{acycl}$ with completeness and soundness properties stated below:
\begin{itemize}
\item (Completeness) if $\phi$ is satisfied by model $\mathcal{M}$ such that $G_\mathcal{M}$ is acyclic, then $\phi\wedge\phi_{acycl}$ is satisfiable.
\item (Soundness) if $\phi\wedge\phi_{acycl}$ is satisfied by model $\mathcal{M}$, then $G_{\mathcal{M}}$ is acyclic.
\end{itemize}
\end{definition}

Analogous to Definition 1, we can define encoding of \emph{s\textnormal{-}t\textnormal{-}reachability}, \emph{s\textnormal{-}t\textnormal{-}unreachability}, and \emph{s\textnormal{-}t\textnormal{-}eventual\textnormal{-}reachability} for $s,t\in V$. The \emph{s\textnormal{-}t\textnormal{-}eventual\textnormal{-}reachability} holds iff $t$ is reachable from all nodes reachable from $s$.

\subsection{Vertex Elimination Graphs}
The concept of vertex elimination graph has originally been introduced in~\cite{ve:ve}. Let $G=(V,E)$ be a directed graph, $G^+=(V,E^+)$ be the transitive closure of $G$, and $O=v_1,...,v_{|V|}$ be any ordering of members of $V$. We construct a sequence of graphs $G_0=G,...,G_{|V|}$ by eliminating vertices of $G$ according to ordering $O$. For each $i>0$, $G_i$ is obtained from $G_{i-1}$, by removing $v_i$, and adding edges from all its in-neighbors to all its out-neighbors. Formally, $G_i=(V_i,E_i)$ is constructed from $G_{i-1}=(V_{i-1},E_{i-1})$ so that $V_i=V_{i-1}\backslash\{ v_i\}$, and $E_i=E_{i-1}\backslash (\{(v_j,v_i)|(v_j,v_i)\in E_{i-1}\}\cup \{(v_i,v_k)|(v_i,v_k)\in E_{i-1}\}) \bigcup D_i$, where $D_i=\{(v_j,v_k)|(v_j,v_i)\in E_{i-1}, (v_i,v_k)\in E_{i-1}, j\ne k\}$. The vertex elimination graph of $G$ according to elimination ordering $O$ is $G^*=(V,E^*)$, where:
\begin{align}
&E^*=\bigcup\limits_{i=0}^{|V|} E_{i}\label{Estar}
\end{align}

The directed elimination width~\cite{width} of ordering $O$ for graph $G$ is defined by the maximum over number of out-neighbors of $v_i$ in $G_i$ for $i=1,...|V|$. The directed elimination width of $G$ is the minimum width over all directed elimination orderings for $G$.

We define $\Delta$ as the set of all triangles produced by elimination ordering $O$ for graph $G$. Members of $\Delta$ are all ordered triples $(v_i,v_j,v_k)$ such that $(v_i,v_k)$ is a member of $D_j$.

Clearly, for each $i$ there is an edge $(v_j,v_k)\in E_i$ only if there is a path in $E_{i-1}$ with length at most 2 from $v_j$ to $v_k$. Therefore, if there is an edge $(v_j,v_k)$ in $E^*$, there must exist a path in $G$ from $v_i$ to $v_j$. We can conclude that $G^*$ is a subgraph of $G^+$. However, the difference between $|E^+|$ and $|E^*|$ depends both on the sparsity of $G$, and the elimination ordering. It has been shown that the problem of finding the optimal ordering, i.e., the ordering that results in the smallest number of edges in the vertex elimination graph, is NP-complete ~\cite{ve:ve}. Nevertheless, there are effective heuristics for finding empirically usable orderings. An examples is \emph{minimum fill-in} heuristic, which chooses $v_i$ so that elimination of $v_i$ adds the minimum number of edges to $G_{i-1}$. Another examples is \emph{minimum degree} heuristic, which chooses $v_i$ with the minimum degree from $G_{i-1}$.


\begin{figure}
  \begin{subfigure}[b]{0.4\columnwidth}
    \includegraphics[width=\linewidth]{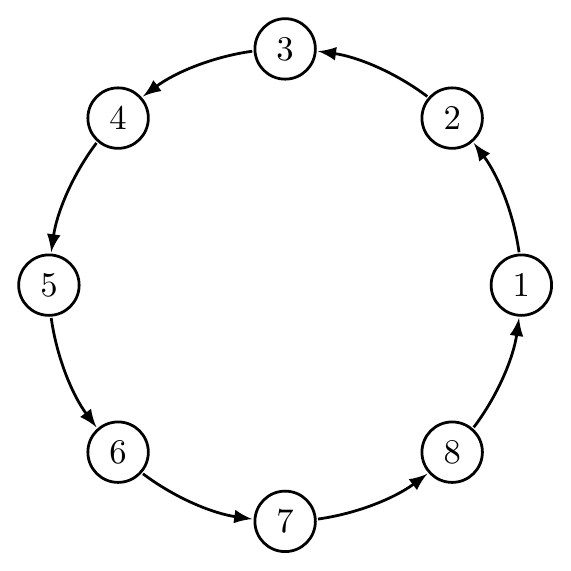}
    \caption{}
    \label{fig:1a}
  \end{subfigure}
  \hfill 
  \begin{subfigure}[b]{0.4\columnwidth}
    \includegraphics[width=\linewidth]{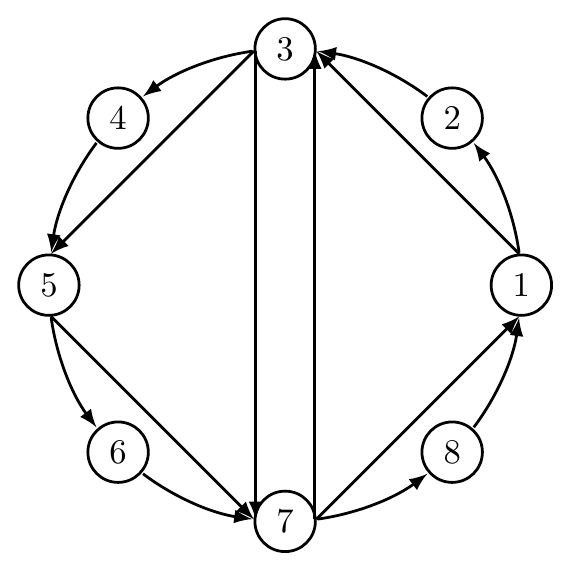}
    \caption{}
    \label{fig:1b}
  \end{subfigure}
  \caption{(a) A simple directed cycle      (b) The vertex elimination graph}
\end{figure}

\begin{example}[Vertex elimination graphs]
Consider $G$ to be the graph depicted in Figure 1(a). There are several elimination orderings that can result in a vertex elimination graph depicted in Figure 1(b), among which one possible ordering is 2,4,6,8,1,5,3,7. The elimination width of this order is 1. Note that, no matter what the elimination ordering is, the vertex elimination graph of graph depicted in Figure 1(a) will have six edges in addition to the edges of $G$ and the elimination width remains 1. That is because for a simple directed cycle, elimination of any node adds one edge. After the elimination of one node, the resulting graph will still be a simple cycle, unless the produced graph has only one vertex. For graphs with number of nodes less than three, no edge can be added by vertex elimination.
\end{example}
\section{Background}
We now explain the methods that have already been introduced for checking acyclicity and reachability when propositional formulas are considered. 
\subsection{Acyclicity}
Various methods have been introduced to explicitly encode acyclicity for symbolic structures with underlying graphs. Examples of general approaches are transitive closure~\cite{tc2,tc3,tc4}, topological sorting with indices~\cite{GebserJR20}, and tree reduction~\cite{tr1,tr2}.
Another approach is to take acyclicity into account when checking the satisfiability of the given formula. This approach, which does not require adding extra clauses to the formula, has been used in GraphSAT~\cite{graphsat} .

Of the methods that use explicit acyclicity encoding, transitive closure and tree reduction are known for better propagation properties~\cite{graphsat}. Since these two methods have been used for our empirical study, we provide more details on them. We also briefly explain GraphSAT which is the current state-of-the-art method for checking acyclicity. We assume that formula $\phi$ with underlying graph $G=(V,E)$ is given.
\subsubsection{Transitive Closure} This encoding of acyclicity for $\phi$, denoted by $\phi_{acycl}^{tc}$, can be produced by conjunction of formulas (\ref{tc:1}) and (\ref{tc:2}):
\begin{align}
 \bigwedge_{(v_i,v_j)\in E, v_k\in V} &e_{i,j} \wedge t_{j,k} \to t_{i,k}\label{tc:1}  \\
 \bigwedge_{(v_i,v_j)\in E} &e_{i,j} \to \neg t_{j,i}\label{tc:2}
\end{align}

Formula (\ref{tc:1}) maintains transitivity, while formula (\ref{tc:2}) ensures acyclicity. This encoding uses $\mathcal{O}(|V|^2)$ variables and produces $\mathcal{O}(|V||E|)$ clauses. If we assume that the graph of Figure 1(a) is the underlying graph of $\phi$, 72 clauses and 64 variables will be used in addition to variables and clauses of $\phi$ in order to encode acyclicity using transitive closure encoding.
\subsubsection{Tree Reduction} This encoding of acyclicity for $\phi$, denoted by $\phi_{acycl}^{tr}$, can be produced by conjunction of formulas (\ref{tr:1}) to (\ref{tr:4}):
\begin{align}
 \bigwedge_{v_i\in V} \bigvee_{n\in \{0,...,|V|-1\}} &d_{i,n}\label{tr:1} \\
 \bigwedge_{v_i\in V, (v_i,v_j)\in E} &e_{i,j} \to \neg d_{i,0}\label{tr:2}  \\
 \bigwedge_{v_i\in V, n\in \{0,...,|V|-2\}} &d_{i,n} \to d_{i,n+1}\label{tr:3}  \\
 \bigwedge_{v_i\in V, (v_i,v_j)\in E, n\in \{1,...,|V|-1\}} &d_{i,n} \wedge e_{i,j}\to d_{j,n-1} \label{tr:4}
\end{align}

Tree reduction encoding is based on the observation that for any acyclic directed graph, every node has a well-defined longest path to a leaf node. For node $v_i$, setting $d_{i,n}$ to $true$ means that k is a lower bound on the length of the longest path from $v_i$ to a leaf node. Formula (\ref{tr:1}) ensures that such lower bounds are assigned to every node. Formula (\ref{tr:2}) guarantees that zero lower bound is given only to the nodes for which all outgoing edges are disabled. Formula (\ref{tr:3}) represents the proper ordering on lower bounds. Finally, formula (\ref{tr:4}) ensures that lower bounds are monotonic along enabled paths, thus ensuring acyclicity.

Similar to the transitive closure encoding, this encoding uses $\mathcal{O}(|V|^2)$ variables and produces $\mathcal{O}(|V||E|)$ clauses. If we assume that the graph of Figure 1(a) is the underlying graph of $\phi$, 128 clauses and 64 variables will be used in addition to variables and clauses of $\phi$ in order to encode acyclicity using tree reduction encoding.

\subsubsection{GraphSAT} Given a mapping of arcs to variables, GraphSAT employs a specialized algorithm for detecting a cycle in the graph induced by those arcs that map to true variables, and to infer that a variable must be false to prevent a cycle emerging in the graph. This algorithm is run together with the unit propagation algorithm inside a standard CDCL implementation. GraphSAT has been shown to outperform explicit SAT encodings for acyclicity as well as non-Boolean representations in terms of linear arithmetic constraints in the  SAT Modulo Theories framework \cite{graphsat}.

\subsection{Reachability}
Checking s-t-reachability as SAT has been studied before. Here we survey three main approaches: explicit encoding, reachability via acyclicity, and implicit reachability checking using GraphSAT.

\subsubsection{Explicit Encoding} Checking unreachability can be done by adding additional formulas to $\phi$ ~\cite{chat,PandeyRintanen18}. Let $s=v_s$ and $t=v_t$ be members of $V$. Encoding of s-t-unreachability for $\phi$, denoted by $\phi_{s\textnormal{-}t\textnormal{-}unreach}$, can be produced by conjunction of formulas (\ref{ur:1}) and (\ref{ur:2}).
\begin{align}
r_{s,s}\wedge \bigwedge_{(v_i,v_j)\in E} e_{i,j} \wedge r_{s,i} \to r_{s,j}\label{ur:1}\\
\neg r_{s,t}\label{ur:2}
\end{align}
Checking reachability, on the other hand, is not as easy as checking unreachability. Encoding of s-t-reachability for $\phi$, denoted by $\phi^{exp}_{s\textnormal{-}t\textnormal{-}reach}$, can be produced by conjunction of formulas (\ref{r:1}) to (\ref{r:3}).
\begin{align}
&r^0_{t,t}\wedge\bigwedge_{v_i\in V \backslash \{t\}}\neg r^0_{i,t} \label{r:1}\\
\bigwedge_{(v_i,v_j)\in E, n=1,...|V|-1} &r^n_{i,t}\to r^{n-1}_{i,t} \vee (e_{i,j}\wedge r^{n-1}_{j,t})\label{r:2}\\
&r^{|V|-1}_{s,t}\label{r:3}
\end{align}

This encoding is derived from ~\cite{PandeyRintanen18}. Setting the variable $r^n_{i,t}$ to $true$ means that there is a path with length at most $n$ from $v_i$ to $t$. The encoding is based on the observation that if $t$ is reachable from $s$, it is reachable by a path with length at most $|V|-1$. Formula (\ref{r:1}) ensures that there is a path from $v_i$ to $t$ with length zero iff $v_i=t$. Formula (\ref{r:2}) guarantees that if there exists a path with length at most $n$ from to $v_i$ to $t$, then there must exist a path with length at most $n-1$ from an out-neighbor of $v_i$ to $t$. Finally, Formula (\ref{r:3}) ensures that there is a path with length at most $|V|-1$ from $s$ to $t$.

This encoding uses $\mathcal{O}(|V|^2)$ variables and produces $\mathcal{O}(|V||E|)$ clauses. If we assume that the graph of Figure 1(a) is the underlying graph of $\phi$, for any $s$ and $t$, 121 clauses and 64 variables will be used in addition to variables and clauses of $\phi$ in order to encode s-t-reachability.

\subsubsection{Reachability by Acyclicity} This encoding has been derived from ~\cite{PandeyRintanen18}. Let $s=v_s$ and $t=v_t$ be members of $V$. Encoding of s-t-reachability by acyclicity for $\phi$, denoted by $\phi^{acycl}_{s\textnormal{-}t\textnormal{-}reach}$, can be produced by conjunction of $\phi'$ and $\phi'_{acycl}$, where $\phi'$ is the conjunction of formulas (\ref{ra:1}) to (\ref{ra:5}), and $\phi'_{acycl}$ is the encoding of acyclicity for $\phi'$, assuming that the underlying graph of $\phi'$ is represented by variables $e'_{i,j}$.

\begin{align}
\bigwedge_{(v_i,v_j)\in E} &e'_{i,j}\to e_{i,j}\label{ra:1}\\
\bigwedge_{(v_i,v_j)\in E} &e'_{i,j} \to r_{i,t}\label{ra:2}\\
\bigwedge_{v_j\in V\backslash\{t\}} &r_{i,t} \to \bigvee_{(v_i,v_j)\in E} e'_{i,j}\label{ra:3}\\
&r_{t,t}\label{ra:4}\\
&r_{s,t}\label{ra:5}
\end{align}

Variables $e'_{i,j}$ are used to represent $G'$, a subgraph of $G$ that is an acyclic directed graph including node $s$, in which all nodes have a path to $v$. Variable $r_{i,t}$ are used to represent the reachability of $t$ from $v_i$. Formula (\ref{ra:1}) ensures that $G'$ is a subgraph of $G$. Formula (\ref{ra:2}) guarantees that every edge of $G'$ goes to a reachable node. Formula (\ref{ra:3}) ensures that all reachable nodes other than $v$ itself have at least one reachable out-neighbor. Formula (\ref{ra:4}) provides the reachability of $t$ from itself, and formula (\ref{ra:5}) guarantees the reachability of $t$ from $s$. 

The acyclicity of $G'$ is necessary because otherwise nodes can obtain reachability from one another in a cycle, without actually having a path to $t$.

This encoding uses $\mathcal{O}(|E|)$ variables and produces $\mathcal{O}(|V|+|E|)$ clauses plus the variables and clauses needed for encoding acyclicity of $\phi'$. Note that if GraphSAT is used for checking acyclicity, then $\phi'_{acycl}$ can be omitted, as acyclicity will be taken into account by GraphSAT while solving the formula. If we assume that the graph of Figure 1(a) is the underlying graph of $\phi$, when taking GraphSAT as the SAT solver, 25 clauses and 16 variables will be used in addition to variables and clauses of $\phi$ in order to encode s-t-reachability by acyclicity.

\subsubsection{Reachability in GraphSAT} As it was mentioned for the case of acyclicity, GraphSAT receives a description of the underlying graph in the input. GraphSAT also admits reachability and non-reachability constraints in its input. While searching for a model, GraphSAT persistently checks that the enabled edges conform to these constraints.

\subsubsection{Eventual Reachability}
Both explicit encoding of reachability and encoding of reachability by acyclicity can be modified to produce encodings for s-t-eventual-reachability. The explicit encoding of s-t-eventual-reachability, denoted by $\phi^{exp}_{s\textnormal{-}t\textnormal{-}event}$, can be produced by conjunction of formulas (\ref{ur:1}), (\ref{r:1}), (\ref{r:2}), and (\ref{er:1}).

\begin{align}
\bigwedge_{v_i\in V} &r_{s,i} \to r^{|V|-1}_{i,t}\label{er:1}
\end{align}

Encoding of s-t-eventual-reachability by acyclicity for $\phi$, denoted by $\phi^{acycl}_{s\textnormal{-}t\textnormal{-}event}$, can be produced by conjunction of $\phi''$, $\phi''_{acycl}$, and (\ref{er:2}), where $\phi''$ is the conjunction of formulas (\ref{ur:1}) and (\ref{ra:1}) to (\ref{ra:4}), and $\phi''_{acycl}$ is the encoding of acyclicity for $\phi''$, assuming that the underlying graph of $\phi''$ is represented by variables $e'_{i,j}$.

\begin{align}
\bigwedge_{v_i\in V} &r_{s,i} \to r_{i,t}\label{er:2}
\end{align}

It is easy to confirm that size properties of the mentioned encodings for s-t-eventual-reachability are the same as their s-t-reachability counterparts.

\section{Encodings with Vertex Elimination Graphs}
Assume that $\phi$ is a propositional formula over the set $X$ of variables, with underlying graph $G=(V,E)$. Let $O$ be an elimination ordering for $G$, $G^*=(V,E^*)$ be the vertex elimination graph of $G$ according to $O$, $\delta$ be the directed elimination width of $O$ for $G$, and $\Delta$ be the set of all triangles produced for $G$ by vertex elimination according to $O$. Also if model $\mathcal{M}$ satisfies $\phi$, let $G_\mathcal{M}$ be the underlying graph of $\mathcal{M}$, and $G^*_\mathcal{M}= (V, E^*_\mathcal{M})$ be the vertex elimination graph of $G_\mathcal{M}$ according to $O$.

\subsection{Encoding of Acyclicity}

The encoding of acyclicity for $\phi$ using vertex elimination according to $O$, denoted by $\phi_{acycl}^{ve}$, is produced by conjunction of formulas (\ref{acve:1}) to (\ref{acve:3}):

\begin{align}
\bigwedge_{(v_i,v_j)\in E} &e_{i,j}\to e'_{i,j}\label{acve:1}\\
\bigwedge_{(v_i,v_j)\in E^*,(v_j,v_i)\in E^*,i < j} &e'_{i,j}\to \neg e'_{j,i}\label{acve:2}\\
\bigwedge_{(v_i,v_j,v_k)\in \Delta} &(e'_{i,j}\wedge e'_{j,k})\to  e'_{i,k}\label{acve:3}
\end{align}

\begin{theorem} [Completeness of $\phi_{acycl}^{ve}$] If $\phi$ is satisfied by any model $\mathcal{M}$ such that $G_\mathcal{M}$ is acyclic, then $\phi\wedge\phi_{acycl}^{ve}$ is satisfiable. \label{complete}
\end{theorem}
\begin{proof} Consider $O'$ to be a topological ordering of members of $V$ according to $G_\mathcal{M}$. We construct valuation function $\mathcal{M}'$ for $\phi\wedge\phi_{acycl}^{ve}$ such that for each $x\in X$, $\mathcal{M}'(x)=\mathcal{M}(x)$, and for each $e'_{i,j}$, $\mathcal{M}'(e'_{i,j})=true$ iff $v_i$ precedes $v_j$ according to $O'$. By definition, $\phi$ is trivially satisfied by $\mathcal{M}'$. Formula (\ref{acve:1}) is satisfied by $\mathcal{M}'$ because if $\mathcal{M}'(e_{i,j})=true$, $v_i$ precedes $v_j$ according to $O'$, thus $\mathcal{M}'(e'_{i,j})=true$. Formula (\ref{acve:2}) is satisfied seeing that if $\mathcal{M}'(e'_{i,j})=true$, then $v_i$ precedes $v_j$, and therefore, $v_j$ cannot precede $v_i$ according to $O'$. Formula (\ref{acve:3}) is satisfied because if $\mathcal{M}'(e'_{i,j})=\mathcal{M}'(e'_{j,k})=true$, then $v_i$ precedes $v_j$ and $v_j$ precedes $v_k$. Therefore, $v_i$ precedes $v_k$ according to $O'$, and $\mathcal{M}'(e'_{i,k})=true$.
\end{proof}

\begin{lemma} If $\phi\wedge\phi_{acycl}^{ve}$ is satisfied by model $\mathcal{M}$, then for every $(v_i,v_j)\in E^*_\mathcal{M}$, we have $\mathcal{M}(e'_{i,j})=true$. \label{meij} 
\end{lemma}
\begin{proof} By formula (\ref{acve:1}) if $\mathcal{M}(e_{i,j})=true$ then $\mathcal{M}(e'_{i,j})=true$. Let $\Delta_\mathcal{M}$ be the set of all triangles produced for $G_\mathcal{M}$ by vertex elimination according to $O$. Since $E^*_\mathcal{M}$ is a subset of $E^*$, we conclude that $\Delta_\mathcal{M}$ is a subset of $\Delta$. The proof is complete by seeing that $\mathcal{M}$ satisfies formula (\ref{acve:3}).
\end{proof}

\begin{lemma} Let $G^*=(V,E^*)$ be a vertex elimination graph of an arbitrary graph $G=(V,E)$ according to an arbitrary elimination ordering $O$. If $G$ has a cycle then for some $v$ and $v'$, we have $(v,v')\in E^*$ and $(v',v)\in E^*$.
\end{lemma}
\begin{proof} We give the proof by induction on the number of vertices in the cycle. Base case: for a cycle of two vertices, the conclusion clearly holds. Induction hypothesis: assume that for $k>2$ and the conclusion holds for any cycle with $k-1$ vertices. For a cycle with $k$ vertices, the cycle has the form $v_0,...,v_{k-1},v_0$. Let $v_i$ be the first vertex in the set $\{v_0,...,v_{k-1}\}$ that is eliminated according to ordering $O$. The edges $(v_{[i-1]_k},v_i)$ and $(v_i,v_{[i+1]_k})$ must be present prior to the elimination of $v_i$. Therefore, $(v_{[i-1]_k},v_{[i+1]_k})$ is a member of $E^*$, constructing a cycle of length $k-1$. The proof is then complete by the induction hypothesis.
\end{proof}

\begin{theorem} [Soundness of $\phi_{acycl}^{ve}$] If $\phi\wedge\phi_{acycl}^{ve}$ is satisfied by model $\mathcal{M}$, then $G_{\mathcal{M}}$ is acyclic.\label{sound}
\end{theorem}

\begin{proof} Assume that $G_{\mathcal{M}}$ has a cycle. Since $G_{\mathcal{M}}$ is a subgraph of $G^*_{\mathcal{M}}$, we conclude that $G^*_{\mathcal{M}}$ has a cycle, too. According to Lemma 2, for some $i$ and $j$, we have: $(v_i,v_j)\in E^*_{\mathcal{M}}$ and $(v_j,v_i)\in E^*_{\mathcal{M}}$. Then, according to Lemma 1, we have $\mathcal{M}(e'_{i,j})=true$ and $\mathcal{M}(e'_{j,i})=true$, which by considering formula (\ref{acve:2}) contradicts the assumption that $\phi\wedge\phi_{acycl}^{ve}$ is satisfied by $\mathcal{M}$.
\end{proof}

Theorem \ref{complete} and Theorem \ref{sound} show that $\phi_{acycl}^{ve}$ is an encoding of acyclicity for $\phi$.

For analyzing the size of $\phi_{acycl}^{ve}$, note that the number of variables in $\phi_{acycl}^{ve}$ is proportional to number of edges in $G^*$, which is $\mathcal{O}(\delta|V|)\subseteq \mathcal{O}(|V|^2)$. This means that in the worst case our vertex elimination based method uses the same asymptomatic number of variables as the transitive closure and tree reduction methods. However, for sparse graphs, directed elimination width can be significantly smaller than $|V|$. By using heuristic methods mentioned in Section 2, one can come up with an ordering with directed elimination width close to that of $G$. For the graph depicted in Figure 1, we need 14 variables in addition to variables in $\phi$, in order to encode acyclicity using vertex elimination encoding, if $O = 2,4,6,8,1,5,3,7$.

The number of clauses in $\phi_{acycl}^{ve}$ is proportional to $|\Delta|+|E^*|$, i. e., the total number of triangles produced by eliminating all vertices plus the number of edges in the vertex elimination graph. When eliminating $v$, the number of triangles produced is at most $in\textnormal{-}degree(v)\times out\textnormal{-}degree(v)\le in\textnormal{-}degree(v)\times \delta$. By summing over all vertices we reach to $\delta|E^*|$, which is $\mathcal{O}(\delta^2|V|)\subseteq\mathcal{O}(|V|^3)$. In sparse graphs $\delta^2$ can be significantly smaller than $|E|$, causing production of smaller number of clauses in comparison with the transitive closure and tree reduction methods. For the graph depicted in Figure 1, we need 15 clauses in addition to clauses in $\phi$, in order to encode acyclicity using vertex elimination encoding, if $O = 2,4,6,8,1,5,3,7$.

\subsection{Encoding of s-t-Reachability}
For encoding of s-t-reachability using vertex elimination according to elimination ordering $O$, we add a restriction on $O$. We demand that $s=v_s$ and $t=v_t$ are ordered after all other vertices by $O$. Assuming this, the encoding of s-t-reachability by using vertex elimination according to elimination ordering $O$, denoted by $\phi_{s\textnormal{-}t\textnormal{-}reach}^{ve}$, is produced by conjunction of formulas (\ref{rve:1}) to (\ref{rve:3}), where $f(e_{i,j})$ is $e_{i,j}$ if $(v_i,v_j)\in E$ and $false$ otherwise.

\begin{align}
\bigwedge_{(v_i,v_j)\in E^*} &e'_{i,j}\to f(e_{i,j})\vee \bigvee_{(v_i,v_k,v_j)\in \Delta} t_{i,k,j}\label{rve:1}\\
\bigwedge_{(v_i,v_k,v_j)\in \Delta} &t_{i,k,j}\to e'_{i,k}\wedge e'_{k,j}\label{rve:2}\\
&e'_{s,t}\label{rve:3}
\end{align}

\begin{theorem} [Completeness of $\phi_{s\textnormal{-}t\textnormal{-}reach}^{ve}$] If $\phi$ is satisfied by any model $\mathcal{M}$ such that $G_\mathcal{M}$ has s-t-reachability, then $\phi\wedge\phi_{s\textnormal{-}t\textnormal{-}reach}^{ve}$ is satisfiable.
\end{theorem}
\begin{proof} Let $\Delta_\mathcal{M}$ be a subset of $\Delta$ constructed similar to $\Delta$ but only by considering edges that are in $G_\mathcal{M}$, i.e., the underlying graph of $\mathcal{M}$. We construct valuation function $\mathcal{M}'$ for $\phi_{s\textnormal{-}t\textnormal{-}reach}^{ve}$ such that for each $x\in X$, $\mathcal{M}'(x)=\mathcal{M}(x)$, for each $e'_{i,j}$, $\mathcal{M}'(e'_{i,j})=true$ iff $(v_i,v_j)\in E^*_\mathcal{M}$, and $\mathcal{M}'(t_{i,k,j})=true$ iff $(i,k,j)\in \Delta_\mathcal{M}$.

Formula (\ref{rve:1}) is satisfied by $\mathcal{M}'$ because for $i$ and $j$ such that $(v_i,v_j)\in E^*$, if $\mathcal{M}'(e'_{i,j})=true$ and either $(v_i,v_j)\notin E$ or $\mathcal{M}'(e_{i,j})=false$, then for some $k$, $(v_i,v_j)$ has been added to $G^*_\mathcal{M}$ when eliminating some $v_k$, and therefore, $\mathcal{M}'(t_{i,k,j})=true$. Formula (\ref{rve:2}) is trivially satisfied by $\mathcal{M}'$. Also, Formula (\ref{rve:1}) is satisfied by seeing that since we have assumed that $v_s$ and $v_t$ are eliminated after all other vertices, if there is a path in $G^*_\mathcal{M}$ from $v_s$ to $v_t$, then $(v_s,v_t)$ must be a member of $E^*_\mathcal{M}$.
\end{proof}

\begin{lemma} If the conjunction of $\phi$, formula (\ref{rve:1}), and formula (\ref{rve:2}) is satisfied by model $\mathcal{M}$, and $\mathcal{M}(e'_{i,j})=true$, then there is a path in $G_\mathcal{M}$ from $v_i$ to $v_j$. 
\end{lemma}

\begin{proof} Without loss of generality assume that vertices are indexed according to elimination ordering $O$. We give the proof by strong induction on $m=min(i,j)$. Base case: for $m=1$, since there are no $(1,k,j)$ or $(i,k,1)$ in $\Delta$, from (\ref{rve:1}) we deduce that there is an edge in $G_\mathcal{M}$ from $e_i$ to $e_j$. Induction hypothesis: assume that for all n such that $1\le n\le m$ and all $i,j\le |V|$, if $n=min(i,j)$ and $\mathcal{M}(e'_{i,j})=true$, then there is a path in $G_\mathcal{M}$ from $e_i$ to $e_j$. We prove that for any $i$ and $j$ such that $min(i,j)=m+1$ and $\mathcal{M}(e'_{i,j})=true$, there is a path in $G_\mathcal{M}$ from $e_i$ to $e_j$. Consider formula (\ref{rve:1}). If $e_{i,j}\in E$ and $\mathcal{M}(e_{i,j})=true$, then conclusion obviously holds. If $e_{i,j}\notin E$ or $\mathcal{M}(e_{i,j})=false$, then there must exist $k$ such that $t_{i,k,j}\in \Delta$ and $\mathcal{M}(t_{i,k,j})=true$. However, in this case since $(v_i,v_j)$ has been added when eliminating $v_k$, $k$ must be smaller than both $i$ and $j$. By formula (\ref{rve:2}), we must have: $\mathcal{M}(e_{i,k})=true$, and $\mathcal{M}(e_{k,j})=true$. Therefore, by induction hypothesis there must be paths from $e_i$ to $e_k$, and from $e_k$ to $e_j$ in $G_\mathcal{M}$. Thus, the conclusion holds.
\end{proof}
\begin{theorem} [Soundness of $\phi_{s\textnormal{-}t\textnormal{-}reach}^{ve}$] If $\phi\wedge\phi_{s\textnormal{-}t\textnormal{-}reach}^{ve}$ is satisfied by model $\mathcal{M}$, then $G_{\mathcal{M}}$ has s-t-reachability.
\end{theorem}
\begin{proof} Since $\mathcal{M}$ satisfies (\ref{rve:3}), by Lemma 3, there must be a path from $v_s$ to $v_t$ in $G_\mathcal{M}$.
\end{proof}

The number of variables used in $\phi_{s\textnormal{-}t\textnormal{-}reach}^{ve}$ is proportional to $|\Delta|+|E^*|$, which in Section 4.1 we showed to be $\mathcal{O}(\delta^2|V|)\subseteq\mathcal{O}(|V|^3)$. The number of clauses is also $\mathcal{O}(\delta^2|V|)$, making this encoding suitable for formulas with sparse underlying graphs. For the graph depicted in Figure 1, we need  28 variables and 27 clauses in addition to clauses and variables of $\phi$, in order to encode 3-7-reachability using vertex elimination encoding, if $O = 2,4,6,8,1,5,3,7$.

\subsection{Encoding of s-t-Eventual-Reachability}
Without loss of generality assume that vertices are indexed according to elimination ordering $O$. We also require $t$ to be ordered after all other vertices by $O$. Assuming these, the encoding of s-t-eventual-reachability by using vertex elimination according to elimination ordering $O$, denoted by $\phi_{s\textnormal{-}t\textnormal{-}event}^{ve}$, is produced by conjunction of formulas (\ref{ur:1}), (\ref{rve:1}), (\ref{rve:2}),  and (\ref{erve:1}).

\begin{align}
\bigwedge_{v_i\in V\backslash\{t\}} r_{s,i}\to \bigvee_{(v_i,v_j)\in E^*,i<j} e'_{i,j}\label{erve:1}
\end{align}

\begin{theorem} [Completeness of $\phi_{s\textnormal{-}t\textnormal{-}event}^{ve}$] If $\phi$ is satisfied by any model $\mathcal{M}$ such that $G_\mathcal{M}$ has s-t-eventual-reachability, then $\phi\wedge\phi_{s \textnormal{-}t\textnormal{-}event}^{ve}$ is satisfiable.
\end{theorem}
\begin{proof} Let $\Delta_\mathcal{M}$ be constructed as it was in the proof of Theorem 3. We construct valuation function $\mathcal{M}'$ for $\phi\wedge\phi_{s\textnormal{-}t\textnormal{-}event}^{ve}$ such that for each $x\in X$, $\mathcal{M}'(x)=\mathcal{M}(x)$, $\mathcal{M}'(e'_{i,j})=true$ iff $(v_i,v_j)\in E^*_\mathcal{M}$, $\mathcal{M}'(t_{i,k,j})=true$ iff $(i,k,j)\in \Delta_\mathcal{M}$, and $\mathcal{M}'(r_{s,i})=true$ iff $v_i$ is reachable from $v_s$ in $G_\mathcal{M}$.

{

\centering
\begin{table*}[ht]

\centering
\begin{tabular}{|c||c|c|c|c||c|c|c|c||c|c|c|}
 \hline

Problem&SAT &$|V|$ &$|E|$ &$\delta$  &\begin{sideways}$\phi_{acycl}^{ve}$\end{sideways}  & \begin{sideways}$\phi_{acycl}^{tr}$\end{sideways} & \begin{sideways}$\phi_{acycl}^{tc}$\end{sideways} &\begin{sideways} {\small GraphSAT}\end{sideways}  & \begin{sideways}$\phi_{s\textnormal{-}t\textnormal{-}event}^{ve}$\end{sideways}  & \begin{sideways}$\phi^{exp}_{s\textnormal{-}t\textnormal{-}event}$\end{sideways} & \begin{sideways} {\small GraphSAT}\end{sideways}  \\
\hline
{comb}&  &  &  &  & &  &  & &  &  &  \\ 
11-2 & F & 286 & 1839 & 8 & \textbf{0.07} & 15.8 & 0.43 & 3.28 & \textbf{0.00} & 49.7 & 3.43  \\
11-3 & T & 428 & 3959 & 11 & \textbf{1.13} & 160 & 2.27 & 4.40 & \textbf{0.02} & --- & 14.7  \\
12-2 & F & 312 & 2008 & 8 & \textbf{0.06} & 38.4 & 0.46 & 3.41 & \textbf{0.00} & 34.6 & 4.01  \\
12-3 & T & 467 & 4323 & 11 & \textbf{0.92} & 35.5 & 1.51 & 5.91 & \textbf{0.03} & --- & 20.3  \\
\hline
{emptycorner}&  &  &  &  & &  &  & &  &  &  \\ 
60-1 & T & 3602 & 17997 & 154 & \textbf{0.53} & 38.3 & 13.9 & 0.56 & 15.5 & --- & \textbf{5.96} \\
65-1 & T & 4227 & 21122 & 181 & 0.75 & 51.3 & 19.4 & \textbf{0.69} & 18.6 & --- & \textbf{8.25} \\
70-1 & T & 4902 & 24497 & 203 & 0.92 & 74.6 & 26.2 & \textbf{0.78} & 21.4 & --- & \textbf{13.0} \\
75-1 & T & 5252 & 26247 & 232 & \textbf{1.00} & 84.8 & 29.9 & 1.03 & 15.9 & --- & \textbf{15.1} \\
\hline
{emptymiddle}&  &  &  &  & &  &  & &  &  &  \\ 
15-2 & T & 452 & 3810 & 70 & \textbf{0.17} & 24.1 & 4.70 & 0.27 & \textbf{1.52} & --- & 114   \\
20-2 & T & 802 & 6875 & 92 & \textbf{2.47} & 65.4 & 6.95 & 5.39 & \textbf{1.92} & --- & 150   \\
25-2 & T & 1252 & 10610 & 118 & \textbf{0.26} & 421 & 9.90 & 7.57 & \textbf{2.93} & --- & 512  \\
30-2 & T & 1802 & 15285 & 158 & \textbf{0.52} & 1465 & 12.9 & 41.8 & \textbf{115} & --- & 1406  \\
\hline
{roomchain}&  &  &  &  & &  &  & &  &  &  \\ 
5-2 & F & 510 & 4223 & 14 & \textbf{43.9} & --- & 535 & --- & \textbf{0.13} & 1441 & 24.8  \\
5-3 & T & 764 & 9183 & 21 & \textbf{27.6} & 699 & 731 & 139 & \textbf{12.4} & --- & 841  \\
6-2 & F & 612 & 5070 & 14 & \textbf{52.8} & --- & 550 & --- & \textbf{0.27} & --- & 60.1  \\
6-3 & T & 917 & 11025 & 21 & \textbf{8.36} & 97.1 & 92.0 & 25.2 & \textbf{1.20} & --- & 676  \\
\hline
{escape}&  &  &  &  & &  &  & &  &  &  \\ 
6-1 & T & 1298 & 26164 & 544 & 11.5 & 16.1 & --- & \textbf{0.08} & 311 & --- & \textbf{5.84}  \\
7-1 & T & 2403 & 50812 & 1003 & 62.8 & 121 & --- & \textbf{0.18} & 865 & --- & \textbf{20.8} \\
8-1 & T & 4098 & 89688 & 1609 & --- & --- & --- & \textbf{0.33}  & --- & --- & \textbf{62.7} \\
9-1 & T & 6563 & 147412 & 2555 & --- & --- & --- & \textbf{0.51} & --- & --- & \textbf{162} \\
\bottomrule
\end{tabular}
\caption{Results of all methods on benchmark problems on planning}
\label{resultstable}
\end{table*}
}

{

\centering
\begin{table}[ht]

\centering
\setlength\tabcolsep{3pt}
\begin{tabular}{|c||c|c|c|c||c|c|c|c|}
 \hline

Size&SAT &$|V|$ &$|E|$ &$\delta$  &\begin{sideways}$\phi_{acycl}^{ve}$\end{sideways}  & \begin{sideways}$\phi_{acycl}^{tr}$\end{sideways} & \begin{sideways}$\phi_{acycl}^{tc}$\end{sideways} &\begin{sideways} {\small GraphSAT}\end{sideways}   \\
\hline
$11\times 11$ & F & 121 & 438 & 14 & \textbf{0.47} & 1.44 & 1.01  & 4.18   \\
$12\times 12$  & T & 144 & 526 & 16 & \textbf{1.57} & 12.5 & 5.16  & 6.52   \\
$13\times 13$  & F & 169 & 622 & 17 & \textbf{4.28} & 14.29 & 23.2 & 17.6   \\
$14\times 14$  & T & 196 & 726 & 18 & \textbf{1.00} & 1480 & 380 & 8.14   \\
$15\times 15$  & F & 225 & 838 & 21 & \textbf{88.1} & --- & 158 & 151   \\
$16\times 16$  & T & 256  & 958 & 22 & 1031 & --- & 463 & \textbf{18.0}  \\
$17\times 17$  & F & 289  & 1086 & 23 & --- & --- & --- &  ---  \\
\hline
$5\times 20$  & T & 100 & 348 & 5 & \textbf{0.01} & 1.24 & 0.16  & 2.99 \\
$5\times 41$  & F & 205 & 726 & 5 & \textbf{0.33} & 2.93 & 2.02 & 9.24  \\
$5\times 60$  & T & 300 & 1068 & 5 & \textbf{0.39} & 1451 & 16.2 & 33.2 \\
$5\times 81$  & F & 405 & 1446 & 5 & \textbf{1.89} & 37.3 & 16.1  & 67.7  \\
$5\times 100$  & T & 500 & 1788 & 5 & \textbf{4.47} & --- & 454  & 110 \\

\bottomrule
\end{tabular}
\caption{Results of acyclicity based methods on finding Hamiltonian cycles in grids}
\label{resultstable2}
\end{table}
}

Formulas (\ref{ur:1}) and (\ref{rve:2}) are trivially satisfied by $\mathcal{M}'$. Formula (\ref{rve:1}) is satisfied by $\mathcal{M}'$ by the same argument made in the proof of Theorem 3. If $v_i\in V\backslash \{t\}$ is reachable from $v_s$ in $G_\mathcal{M}$, since $G_\mathcal{M}$ has s-t-eventual-reachability property, there must exist a path from $v_i$ to $t$. Not all nodes in such a path can have indices less than $i$. That is because we have assumed that $O$ puts $t$ after every other vertex. Assume that we traverse the mentioned path until we visit the first node $v_j$ such that $i<j$. Since according to $O$ all nodes before visiting $v_j$ are eliminated before eliminating $v_i$ and $v_j$, we conclude that $(v_i,v_j)\in E^*_{\mathcal{M}}$ and thus, $(v_i,v_j)\in E^*$. Then we have: $\mathcal{M}'(e'_{i,j})=true$. We can conclude that (\ref{erve:1}) is also satisfied by $\mathcal{M}'$.
\end{proof}

\begin{theorem} [Soundness of $\phi_{s\textnormal{-}t\textnormal{-}event}^{ve}$] If $\phi\wedge\phi_{s\textnormal{-}t\textnormal{-}event}^{ve}$ is satisfied by model $\mathcal{M}$, then $G_{\mathcal{M}}$ has s-t-eventual-reachability.
\end{theorem}

\begin{proof} From formula (\ref{ur:1}), we can conclude that if $v_i$ is reachable from $s$, then we have: $\mathcal{M}(r_{s,i})=true$. By formula (\ref{erve:1}), for some $j$ such that $i<j$, we have $\mathcal{M}(e'_{i,j})=true$. By Lemma 3, there must be a path from $v_i$ to $v_j$ in $G_\mathcal{M}$. Therefore, $v_j$ is reachable from $v_s$. We can repeat the same argument and find paths from $s$ to vertices with increasing indices. Because $O$ puts $t$ after every other vertex, such paths must at some point reach $t$.
\end{proof}

It is easy to see that size properties of the encoding of s-t-eventual-reachability by using vertex elimination is asymptotically the same as those of the encoding of s-t-reachability by using vertex eliminations.

\section{Empirical Results and Discussion}

For analyzing our methods empirically, we have used the benchmark problem sets of ~\cite{PandeyRintanen18} that includes a total of 108 satisfiable and unsatisfiable instances with underlying graphs. We have chosen these benchmarks for several reasons. Firstly, we would want to show that vertex elimination based acyclicity and reachability checking methods can have direct impact on current research in AI. Furthermore, problem sets of \cite{PandeyRintanen18} include instances with underlying graphs of diverse characteristics, enabling us to express the potentials and limitations of our vertex elimination based methods. Moreover, these benchmark problems can be solved by both reachability and eventual acyclicity checking, and are accompanied with tools that transform reachability constrained problems to equivalent acyclicity constrained problems. Also, satisfying graph constraints is a critical requirement for solving the mentioned problems. This property is essential when analyzing methods for solving formulas with underlying graphs: if the bottleneck of solving a problem is in the propositional aspect rather than in graph constraints, the effectiveness of graph constraints encoding methods can be overshadowed by the effort made by the solver on figuring out the propositional structure of problems.

We have also tested the acyclicity encodings on checking the existence of Hamiltonian cycles in two dimensional grids of various sizes. Existence of Hamiltonian cycles can easily be encoded into formulas with underlying graphs and constraint of acyclicity ~\cite{graphsat}. We have chosen this benchmark for two main reasons. Firstly, checking acyclicity is a bottleneck to solve problem instances produced by the mentioned encoding. Secondly, the elimination width of grids is the minimum of the dimensions, a parameter that can be controlled easily. A grid of $n\times m$ has a hamiltonian cycle iff at least one of the dimensions is even. We use grids of size $n\times n$ for $n=11,...,17$. We also use $5\times 20$, $5\times 41$, $5\times 60$, $5\times 81$, and $5\times 100$ grids to produce satisfiable and unsatisfiable formulas of increasing size with low elimination width underlying graphs.

We implemented our vertex elimination encodings, as well as the transitive closure and tree reduction methods mentioned in Section 3. As the heuristic for elimination orderings of the vertex elimination methods, we have used \emph{mindegree}, i.e., eliminating a vertex with minimal total number of incoming and outgoing edges in the graph produced after the elimination of previously eliminated vertices. As the SAT solver for methods other than GraphSAT, we have used  \emph{Kissat} ~\cite{kissat}, which has won the first place in the main track of the SAT Competition 2020. We also used GraphSAT, which becomes the Glucose SAT solver \cite{AudemardSimon09} in the absence of special graph constraints in the input formula, as the solver for other methods. According to our experiments, when using variable elimination encodings, Glucose outperforms GraphSAT with special graph constraints in almost every case in which Kissat does, although in some cases with a smaller margin. Glucose even has a better performance than Kissat on some of the instances. However, we only present the results of \emph{Kissat}, which were in general more favorable, for methods other than GraphSAT. All experiments were run on a cluster of Linux machines, using a timeout of 1800 seconds per instance, and a memory limit of 64 GB.

Table \ref{resultstable} shows the results of different methods on a few problems that are hardest to solve in each problem set of ~\cite{PandeyRintanen18}. The table has three parts, separated by double lines. The first part is dedicated to the characteristics of the underlying graph of each instance. This part include number of vertices, $|V|$, number of edges, $|E|$, and the elimination width of the elimination ordering used by vertex elimination methods, $\delta$. Note that since for reachability checking we require the target vertex to be ordered after all other vertices,  the elimination width for reachability checking by vertex elimination can be different from that of the acyclicity checking method. However, in our experiments we observed that the difference between these two widths is very small (often zero) for the instances under study. Therefore, we only present the elimination width of the acyclicity checking method in Table \ref{resultstable}.

The second part of Table \ref{resultstable} shows the results for solving instances by acyclicity checking, while the third part includes the corresponding results for reachability checking. In the case that the instance cannot be solved either because of time or memory constraints, no result has been reported.

\begin{figure}[ht]

    \includegraphics[width=0.95\linewidth]{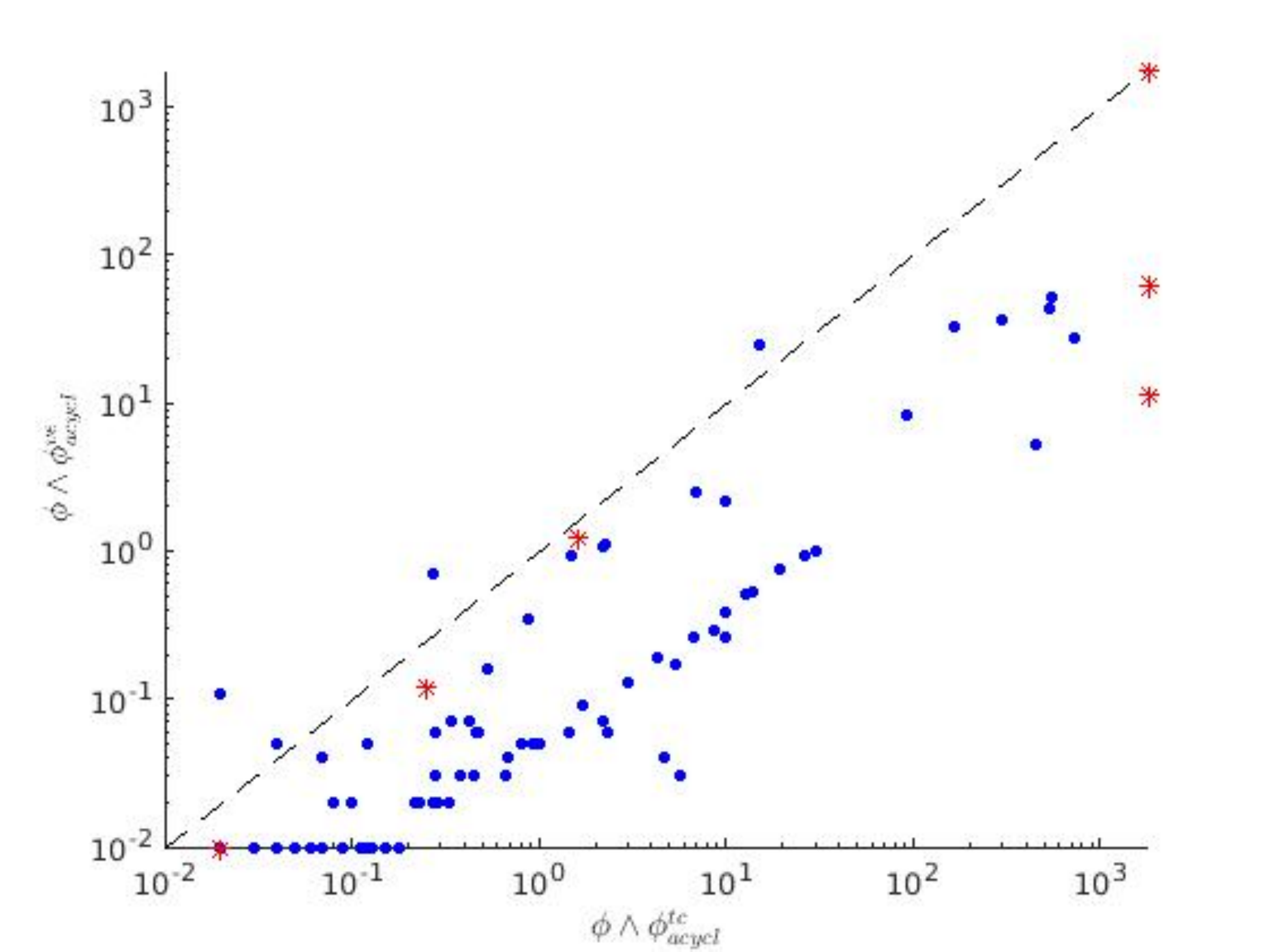}
  \caption{Time (in seconds) needed to solve instances produced by using $\phi^{ve}_{acycl}$ and $\phi^{tc}_{acycl}$ for all instances of (Pandey and Rintanen, 2018).}
      \label{fig:2}
\end{figure}

\begin{figure}[ht]

    \includegraphics[width=\linewidth]{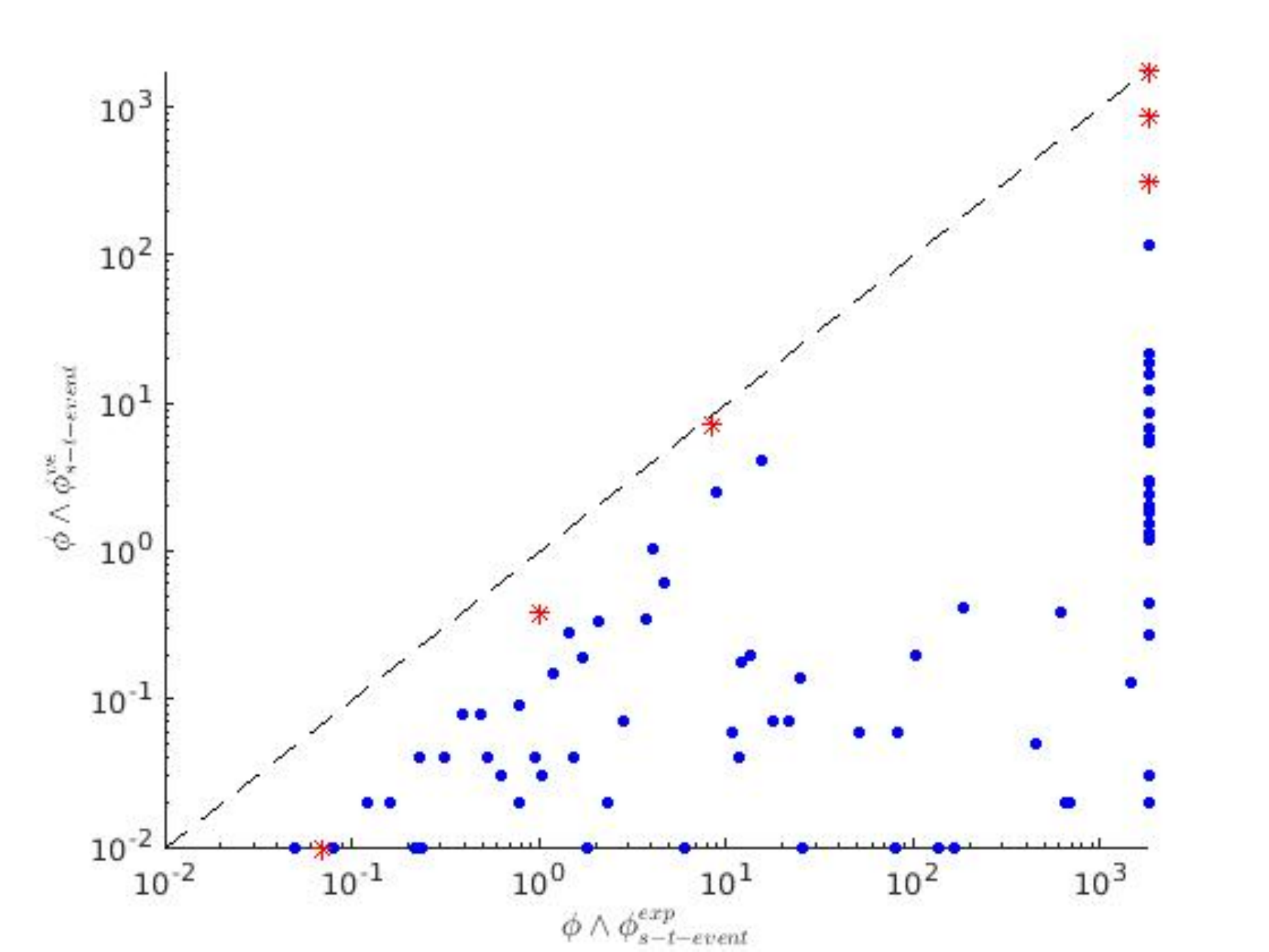}
    
  \caption{Time (in seconds) needed to solve instances produced by using $\phi^{ve}_{s\textnormal{-}t\textnormal{-}event}$ and $\phi^{exp}_{s\textnormal{-}t\textnormal{-}event}$ for all instances of (Pandey and Rintanen, 2018).}
  \label{fig:3}
\end{figure}

\begin{figure}[ht]

    \includegraphics[width=\linewidth]{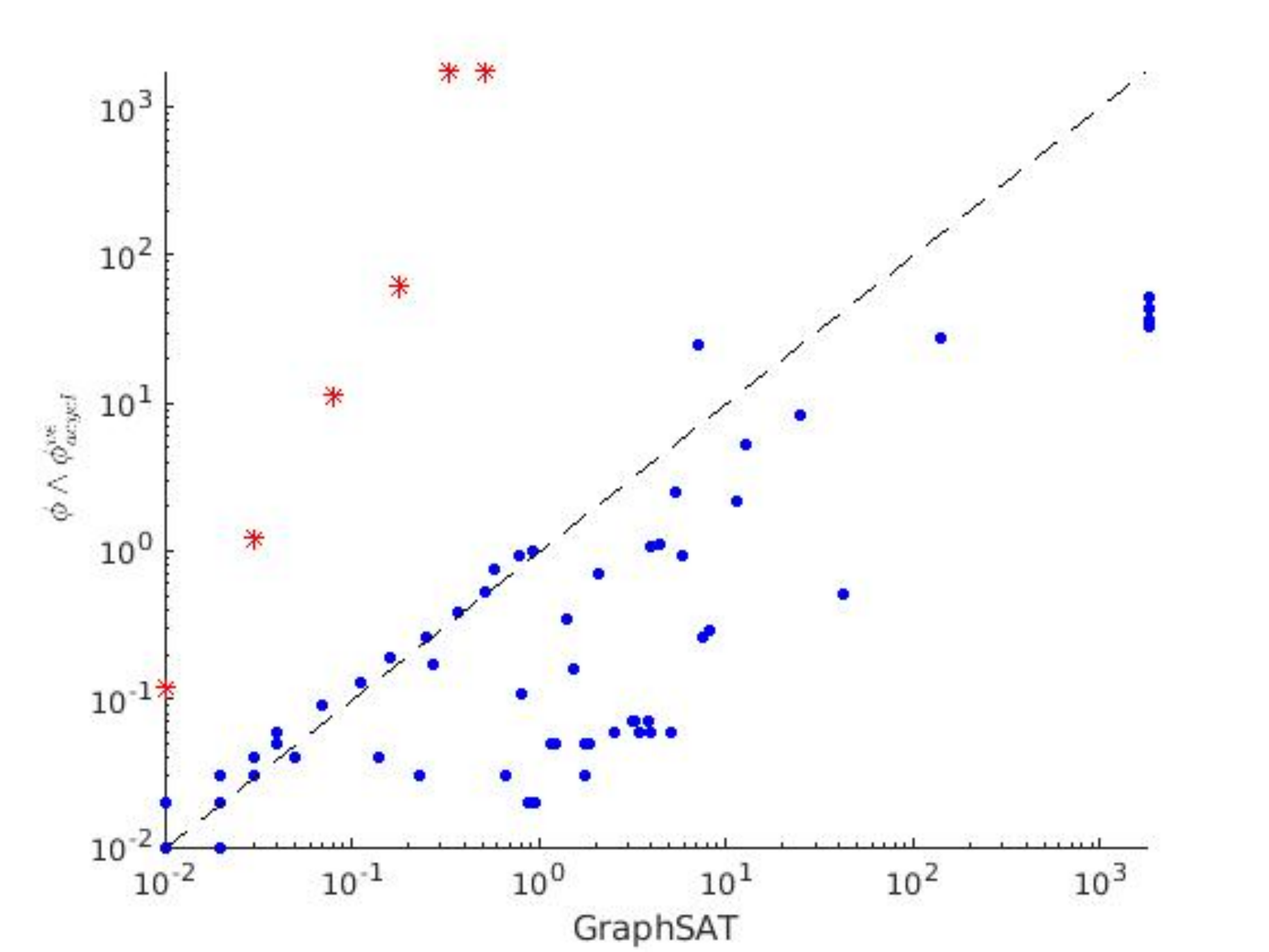}
  \caption{Time (in seconds) needed by GraphSAT versus time needed for Kissat when given $\phi\wedge\phi^{ve}_{acycl}$ for all instances of  (Pandey and Rintanen, 2018).}
      \label{fig:4}
\end{figure}

\begin{figure}[ht]

    \includegraphics[width=\linewidth]{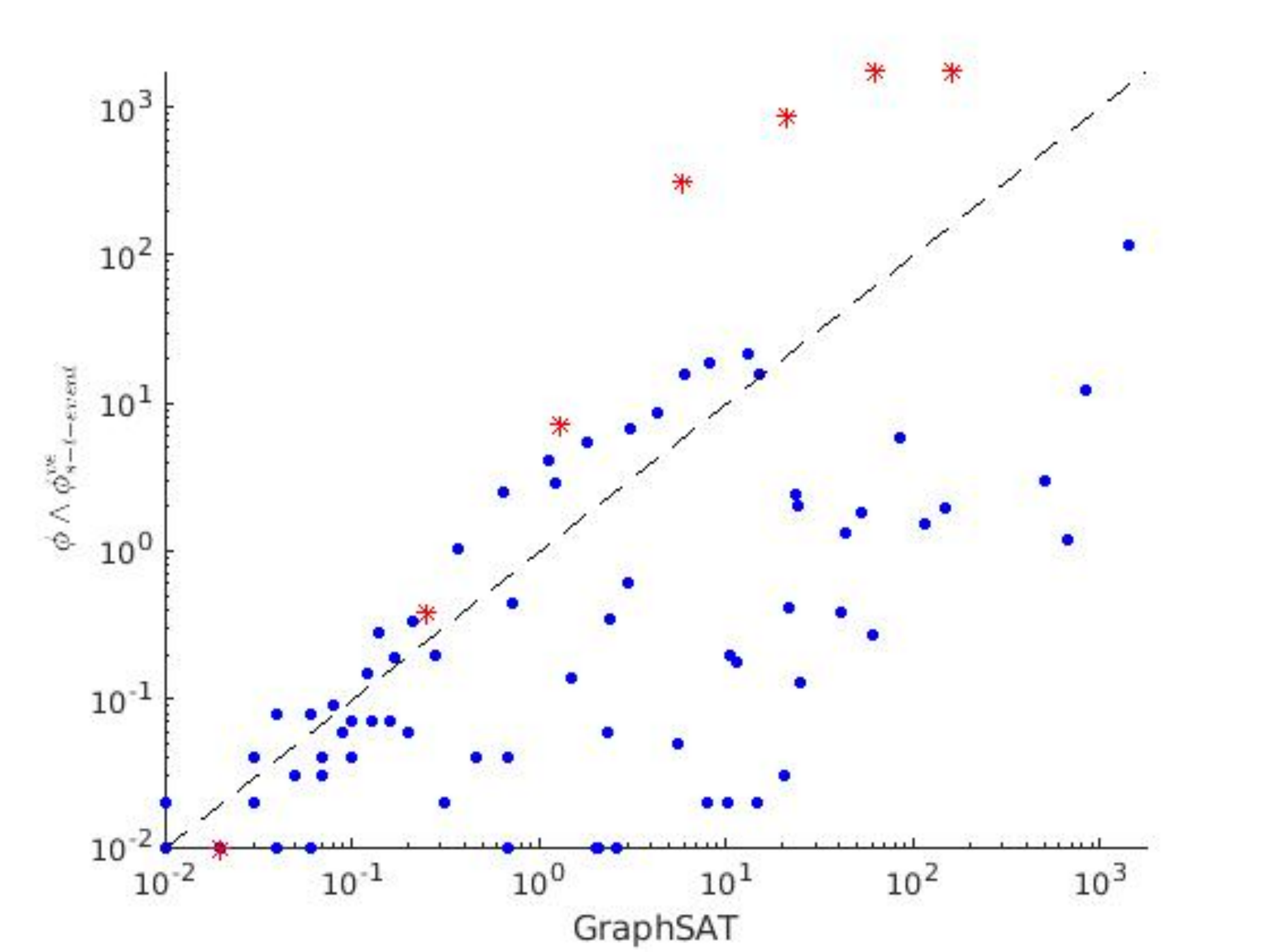}
    
  \caption{Time (in seconds) needed by GraphSAT versus time needed for Kissat when given $\phi\wedge\phi^{ve}_{s\textnormal{-}t\textnormal{-}event}$ for all instances of  (Pandey and Rintanen, 2018).}
  \label{fig:5}
\end{figure}

As it can be seen in Table \ref{resultstable}, the elimination width is quite small for \emph{comb} and \emph{roomchain} problem sets. In fact, the same elimination widths have been observed for all problems of these two problem sets. For \emph{emptycorner} and \emph{emptymiddle}, the elimination width grows as problem size increases. However the elimination width is small compared to the number of indices in these two problem sets. We regard instances from \emph{comb}, \emph{emptycorner}, \emph{emptymiddle}, and \emph{roomchain} as instances with sparse underlying graphs.  The problem set \emph{escape}, on the other hand, is not sparse at all. The elimination width for \emph{escape9-1} is as high as 39 percent of the number of vertices. We do not expect our vertex elimination methods to scale well for \emph{escape} problem set. We distinguish the results for problems of \emph{escape} problem set in the scatter plots of Figure \ref{fig:2} to Figure \ref{fig:5} described below with ``*'' symbols of larger size compared to that of other instances.

The results for Hamiltonian cycle detection are presented in Table \ref{resultstable2}.

\subsection{Vertex Elimination Versus Other Explicit Encodings}
Our vertex elimination based methods are considered to be explicit in the sense that they incorporate graph constraints into the encoding. Therefore, it would be interesting to see how these methods compare with other explicit methods.

From Table \ref{resultstable} it can be observed that for the hardest problems, encoding acyclicity using vertex elimination significantly outperforms transitive closure and tree reduction methods. We have also presented the comparison between acyclicity using vertex elimination and transitive closure on all instances of ~\cite{PandeyRintanen18} in Figure \ref{fig:2}. Since transitive closure outperforms tree reduction for almost all of our instances, we do not present the corresponding comparison between acyclicity using vertex elimination and tree reduction.

Table \ref{resultstable} also shows that encoding eventual reachability using vertex elimination significantly outperforms the explicit encoding of eventual reachability mentioned in Section 3. Figure \ref{fig:3} presents the results of these two methods for all instances from ~\cite{PandeyRintanen18}.

Figure \ref{fig:2} and Figure \ref{fig:3} show that our vertex elimination based methods significantly outperform other explicit encoding methods on the benchmark problems of ~\cite{PandeyRintanen18}, even for the instances with dense underlying graphs. These results are important because explicit encodings allow using off-the-shelf state-of-the-art SAT solvers without any necessity for modifying the solver.

Considering the Hamiltonian cycles detection problem, as it can be seen in Table \ref{resultstable2}, the vertex elimination based encoding outperforms other explicit encodings, as long as the elimination width is small. However, the performance gain over transitive closure encoding diminishes rapidly as the elimination width grows.

\subsection{Vertex Elimination Versus GraphSAT}

Table \ref{resultstable} shows that our encodings of acyclicity and eventual reachability using vertex elimination outperform GraphSAT in the hardest problems of \emph{comb}, \emph{emptymiddle}, and \emph{roomchain} problem sets. For \emph{emptycorner} problem set, when encoded by acyclicity, the performance of our method and that of GraphSAT are roughly the same. If \emph{emptycorner} is encoded by eventual reachability, GraphSAT outperforms our method. GraphSAT also significantly outperforms our methods in \emph{escape} problem set. Figure \ref{fig:4} and Figure \ref{fig:5} show how our acyclicity and reachability checking methods compare with GraphSAT on all instances, respectively. We have omitted the instances that are solved by both methods in less than 10 milliseconds.

From Figure \ref{fig:4} and Figure \ref{fig:5} it can be observed that, as far as sparse instances are considered, the benefits of using vertex elimination outweigh its disadvantages for the problem sets under study. GraphSAT heavily outperforms our methods in instances with dense underlying graphs. Nevertheless, the elimination width of any elimination order can be computed beforehand in polynomial time. In other words, one can use a preprocessing method to check whether vertex elimination based encodings can be considered promising or not.

Even if the graph is sparse, there is no guarantee that vertex elimination based methods outperform GraphSAT. Note that we are addressing the problem of checking the satisfiability of a given formula with an underlying graph, rather than checking whether a given graph has a specific property. GraphSAT takes a \emph{lazy} approach when dealing with formulas with underlying graphs. It waits for the graph constraints to be close to violation, and only then takes action by enforcing the constraints. Our vertex elimination based methods, and also other explicit encodings, are more \emph{eager} by comparison: these methods encode constraints of the underlying graph without taking into account whether the constraint will be violated when solving the problem or not. In other words, it can happen that some constraints encoded by explicit methods never get violated during the search. Similar to eager approaches in other contexts, it is only normal that the counterpart lazy approach would perform better for some problems.

As for Hamiltonian cycle detection problem set, it can be observed from Table \ref{resultstable2} that for smaller elimination widths, our vertex elimination method has a considerably better performance in comparison with GraphSAT. Nevertheless, GraphSAT significantly outperforms our method on the grid of size $16\times 16$, where the elimination width is relatively high.

\section{Conclusion and Future Research}

We have addressed the problem of checking the satisfiability of propositional formulas with underlying graphs, in the presence of acyclicity and reachability constraints. Novel methods that leverage the sparsity of underlying graphs in order to produce compact encodings for the constraints were introduced. We proved soundness and completeness for each method, and also provided theoretical evidence for efficiency of the methods by parameterized complexity analysis based on the elimination width of the elimination orderings. Moreover, we empirically showed that our new methods can outperform GraphSAT and other encoding methods, especially when underlying graphs are sparse.

As mentioned in Section 5, our variable elimination based methods only take into account the structural properties of the underlying graphs. However, the propositions represented by graph edges can have rich semantics based on the problem knowledge represented in the input formula. An interesting direction for future research could be taking advantage of such semantics along with structural properties of the underlying graphs in order to produce still better encodings.

\bibliographystyle{kr}
\bibliography{kr-sample}

\end{document}